\documentclass[runningheads]{llncs}

\usepackage{accv}

\usepackage{accvabbrv}

\usepackage{graphicx}
\usepackage{booktabs}
\usepackage[inline]{enumitem}
\usepackage{comment}
\usepackage{placeins}
\usepackage{multirow}
\usepackage{flushend}
\usepackage{wrapfig} % put this in preamble
\usepackage[accsupp]{axessibility}  % Improves PDF readability for those with disabilities.

\usepackage{hyperref}

\usepackage{orcidlink}

\begin{document}

% ---------------------------------------------------------------
% TODO REVIEW: Replace with your title
\title{SynMulti: Synthetic-to-Real Learning for Multimodal Video Understanding} 

% TODO REVIEW: If the paper title is too long for the running head, you can set
% an abbreviated paper title here. If not, comment out.
\titlerunning{Synthetic-to-Real Learning for Multimodal Video Understanding}

% TODO FINAL: Replace with your author list. 
% Include the authors' OCRID for the camera-ready version, if at all possible.
\author{
Tanzila Rahman$^{1,2}$\thanks{Work done while at University of British Columbia and Vector Institute for AI.} \quad
Renjie Liao$^{1,2,3}$ \quad
Leonid Sigal$^{1,2,3}$\\[0.5em]
$^1$University of British Columbia \quad
$^2$Vector Institute for AI \quad
$^3$Canada CIFAR AI Chair
}

% TODO FINAL: Replace with an abbreviated list of authors.
\authorrunning{T. Rahman et al.}
% First names are abbreviated in the running head.
% If there are more than two authors, 'et al.' is used.

% TODO FINAL: Replace with your institution list.
\institute{}

\maketitle

\begin{abstract}
  Training multimodal large language models (MLLMs) for video understanding requires large-scale annotated data spanning diverse tasks such as object counting, question answering, and segmentation. However, collecting and annotating multimodal video data in real-world is costly, slow, and inherently limited in diversity and coverage. To address this challenge, we propose~\textbf{SynMulti} dataset, along with a unified synthetic data generation pipeline capable of automatically producing unlimited multimodal video data with rich and diverse supervision. Our framework supports multiple task formats within a single pipeline, enabling scalable and consistent data creation across tasks.
To further enhance reasoning ability, we introduce a VQA-based fine-tuning strategy that trains models to answer structured questions about visual content rather than relying solely on captions or simple instructions. This formulation encourages deeper visual grounding and reasoning.
We evaluate our approach in three challenging tasks: video object counting, video-based visual question answering, and video object segmentation. Experimental results demonstrate that models trained predominantly on synthetic data generalize effectively to real-world datasets, often outperforming traditionally trained counterparts. Our findings highlight the potential of unified synthetic data pipelines as a scalable alternative to expensive real-world annotation for multimodal video understanding.
  
\end{abstract}
 
\section{Introduction}
\label{sec:intro}

\begin{figure}[t]
    \centering
    \includegraphics[width=\linewidth]{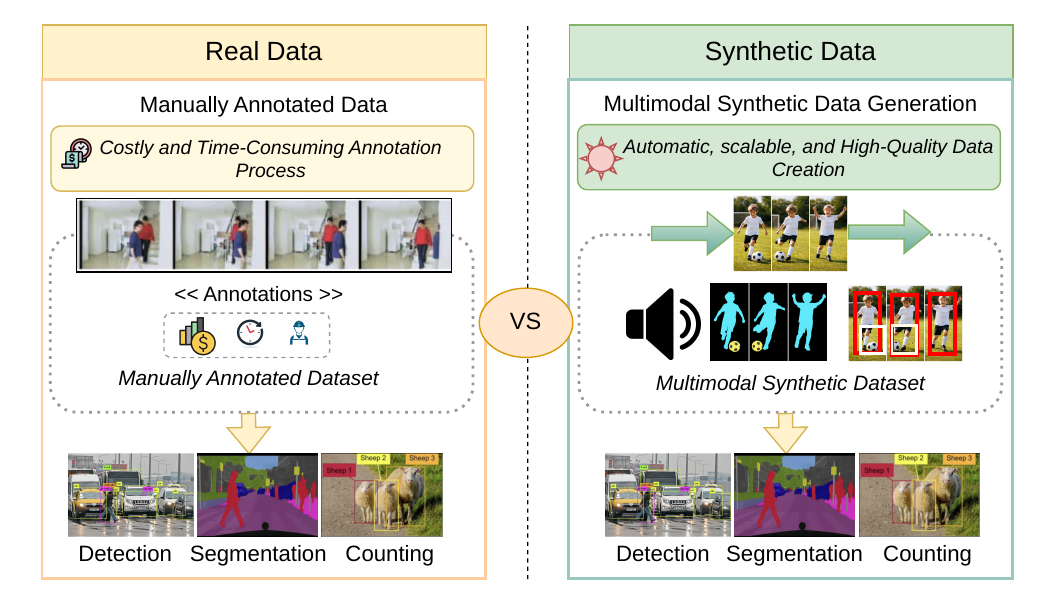}
    \caption{\textbf{Comparison of real and synthetic data for training video foundation models.} Real data (left) requires manual annotation, making it costly and labor-intensive, while synthetic data (right) is automatically generated, offering a scalable and efficient alternative for creating high-quality training datasets. }
    \label{fig:teaser}
    \vspace{-0.32in}
\end{figure}

Multimodal large language models (MLLMs) have demonstrated remarkable progress in understanding and reasoning over visual and textual data. Recent advances in VLM models have enabled a wide range of applications, including image captioning~\cite{nguyen2023improving, hu2023promptcap, hu2022scaling}, visual question answering (VQA)~\cite{jia2025vqa2, tiong2022plug, changpinyo2022all}, and video understanding~\cite{li2025videochat, buch2022revisiting, tang2025video}. These successes are largely driven by the scale of the models and the training data. However, the acquisition of large-scale, high-quality multimodal annotations remains a fundamental bottleneck. This challenge is particularly acute for video-based tasks, which require dense supervision across both spatial and temporal dimensions, including frame-level annotations, object tracking, and temporally consistent labeling~\cite{liu2023revisiting,isobe2020revisiting}. Such requirements significantly increase the cost, complexity, and time associated with dataset construction, making it difficult to scale real-world data collection. Moreover, real-world datasets often suffer from biases, uneven coverage of reasoning patterns, and limited diversity, which restrict the ability of models to learn generalizable multimodal reasoning skills~\cite{ravishankara2025artificial, wang2024exploring}.

Data quality is a critical factor in determining the effectiveness of modern machine learning systems. Although real-world data are ubiquitous, they are often noisy, incomplete, biased, or expensive to annotate. In addition, when sensitive information is involved (e.g., medical or personal data), privacy constraints further limit the availability and usability of data~\cite{de2020overview, xu2021privacy}. These challenges have motivated growing interest in synthetic data, artificially generated data designed to mimic the statistical properties of real data are a powerful alternative to traditional data collection. Unlike real data, synthetic data can be generated at scale with precise control over quality, balance, and diversity while also avoiding privacy concerns. However, for synthetic data to be effective, it must be plausible and capture the underlying distributions and structural patterns of real-world data. Early approaches such as Random OverSampling and SMOTE~\cite{chawla2002smote} attempted to address data imbalance through simple transformations, but their ability to model complex dependencies was limited. More recent advances in deep generative models, including Variational Autoencoders (VAEs)~\cite{diederik2019introduction} and Generative Adversarial Networks (GANs)~\cite{goodfellow2020generative}, have enabled the generation of high-fidelity synthetic data across a wide range of domains.
%, particularly in computer vision.

Despite these advances, most existing multimodal LLMs still rely heavily on real-world datasets with manual annotations in the form of captions, instructions, or aligned image–text pairs~\cite{lin2014microsoft, krishna2017visual, li2023uniformerv2, soomro2012ucf101, xu2018youtube, damen2020epic}. Although effective to some extent, these annotation strategies are limited in diversity and often fail to capture complex spatial–temporal relationships or higher-order reasoning patterns. Furthermore, the availability of such annotated datasets is uneven across tasks, modalities, and domains, creating a major bottleneck for scaling and generalizing multimodal models, especially in video-centric settings.

To address these challenges, we propose a synthetic data generation pipeline that enables the creation of unlimited multimodal data with rich and structured annotations. By leveraging controllable simulation and automated labeling mechanisms, our pipeline produces high-quality synthetic datasets without the need for costly human annotations. This approach allows precise control over object configurations, scene layouts, temporal dynamics, and reasoning complexity, while ensuring balanced coverage across diverse scenarios. The generated data supports a wide range of vision language tasks, including object counting~\cite{pothiraj2025capture, wang2024language, binyamin2025make}, VQA~\cite{antol2015vqa, goyal2017making, singh2019towards}, and object segmentation~\cite{yao2020video, gao2023deep, caelles2017one}, which makes it suitable for training robust and generalizable multimodal models (see Figure~\ref{fig:teaser}).

In addition to the data generation pipeline, we introduce a novel fine-tuning strategy based on Visual Question Answering (VQA), rather than conventional caption-based or instruction-following supervision~\cite{wang2025clipcap, somepalli2024calvin, zhang2024qwen}. Unlike captions, which often emphasize surface-level descriptions, VQA explicitly requires models to attend to specific visual evidence and reason over objects, attributes, and temporal relationships to produce correct answers. By conditioning learning on targeted questions, this paradigm enforces tighter alignment between visual perception and language understanding, encouraging the model to focus on task-relevant regions and frames while suppressing irrelevant context. In addition, VQA naturally supports various forms of reasoning, including counting, comparison, temporal inference, and causal reasoning—capabilities that are critical for complex video understanding tasks.

We evaluate our approach on three challenging video-based tasks: video object counting, video-based VQA, and video object segmentation. Extensive experiments demonstrate that models trained on our synthetic data generalize effectively and achieve competitive or improved performance on real-world benchmarks. This strong transfer performance can be attributed to the structured, controllable, and diverse nature of synthetic supervision, which exposes models to a wide range of object interactions, temporal dynamics, and reasoning patterns that are difficult to obtain at scale from real-world videos. Consequently, the model learns general multimodal reasoning principles rather than overfitting to dataset specific biases, while significantly reducing the cost and effort associated with manual data annotation.

In summary, we make the following key contributions in this paper:
\vspace{-0.1in}
\begin{enumerate}
    \item We introduce~\textbf{SynMulti} dataset, along with a synthetic data generation pipeline capable of producing unlimited data with various types of annotation, including texts, images, videos, audios and so on. This enables training of multimodal large language models (LLMs) with reduced dependency on real-world data.
    
    \item Unlike most existing LLMs, which are fine-tuned using instructions or image captions, we propose a novel fine-tuning approach based on Visual Question Answering (VQA) to enhance the reasoning abilities of multimodal LLMs.
    
    \item We evaluate our approach on three tasks, video object counting, video-based VQA, and video object segmentation; and demonstrate that models trained with our synthetic data show improved performance on real-world benchmarks.
\end{enumerate}
\section{Related work}
\label{sec:related_work}

\subsection{Synthetic Data and Its Evaluation } 
Synthetic data has attracted increasing attention due to its scalability and reduced annotation cost. In safety-critical applications, research has focused primarily on synthetic data detection to prevent the misuse of AI-generated content ~\cite{gragnaniello2021gan,hou2023evading}. Early work explored detection in images and audio ~\cite{barni2020cnn, frank2020leveraging}, while more recent efforts extended to video, such as DuB3D~\cite{ji2024distinguish} and AIGVDet ~\cite{bai2024ai}. Most existing approaches formulate detection as a binary classification problem between authentic and synthetic data, limiting interpretability. Some studies attempt to improve interpretability through latent representation analysis~\cite{dong2022explaining}, feature attributions~\cite{chai2020makes}, or artifact localization~\cite{shao2023detecting}. However, these explanations often remain abstract and poorly aligned with human-understandable reasoning.

To standardize the evaluation, several synthetic detection benchmarks have been proposed. Fake2M~\cite{lu2023seeing}, HC3~\cite{guo2023close}, and ASVSpoof 2019~\cite{wang2020asvspoof} evaluate traditional deepfake detection methods across modalities. More recent benchmarks such as VANE~\cite{gani2025vane} and FakeBench~\cite{li2025fakebench} assess multimodal large models (LMMs) but focus on limited modalities or task types. LOKI~\cite{ye2024loki} introduces a wider multimodal coverage and includes explanation-based evaluation tasks.

Although these works emphasize the detection of  synthetic data, our work takes a complementary perspective: instead of identifying synthetic content, we leverage synthetic generation as a scalable training paradigm for multimodal video understanding. Our unified pipeline generates structured supervision across multiple tasks, enabling both training and standardized evaluation under controlled variations.

\vspace{-0.1in}
\subsection{Compositionality and Reasoning in MLLMs}
Despite strong performance in multimodal tasks~\cite{lu2022learn, yue2024mmmu, bai2024survey}, vision-language models (VLMs) and large multimodal models (LMMs) still struggle with compositional reasoning. Benchmarks such as What’sUp~\cite{kamath2023s}, SPEC~\cite{peng2024synthesize}, ARO~\cite{yuksekgonul2022and}, Winoground~\cite{thrush2022winoground}, SNARE~\cite{wang2023self}, and VL-CheckList~\cite{zhao2022vl} reveal weaknesses in understanding spatial relations, object attributes, counting, and lexical variations. SugarCrepe~\cite{hsieh2023sugarcrepe} and SugarCrepe++~\cite{dumpala2024sugarcrepe++} further show that models often fail to distinguish subtle semantic or lexical variations, suggesting that prior evaluations may overestimate compositional ability. To improve compositionality, prior methods leverage dense captions~\cite{doveh2023dense, khan2024figclip}, distill knowledge from pretrained models such as SDS-CLIP~\cite{basu2023augmenting}, SF-CLIP~\cite{sameni2024building}, and IL-CLIP~\cite{zheng2024iterated}, incorporate structural supervision through scene graphs~\cite{yellinek20233vl, herzig2023incorporating, huang2024structure}, generate synthetic negatives using rule-based tools or generative models~\cite{cascante2024natural, castro2024clove, doveh2023teaching, zhang2024countercurate, zhang2024contrasting}, or enforce fine-grained alignment constraints~\cite{zhang2024contrasting, bica2024improving, kim2023misalign}. However, these approaches often depend on expensive annotations, inherit pretrained limitations, or suffer from inconsistent alignment quality in synthetic data.

In contrast, our unified synthetic pipeline directly generates multimodal data with structured supervision, including question–answer pairs, object counts, and segmentation masks. By training models through VQA-style reasoning rather than simple caption matching, we encourage stronger compositional grounding across spatial and temporal dimensions.

\vspace{-0.1in}
\subsection{Synthetic Data for Multimodal Learning}
Synthetic data has long been used in simulation-based training across domains~\cite{he2022generate, kumar2020data, meng2022generating, rosenberg2019speech}. Graphics engines and simulators enable the generation of scalable datasets~\cite{varol2021synthetic, peng2017visda}, but the resulting distributions often diverge from real-world data. Recent advances in generative models have narrowed this gap, enabling higher-fidelity synthetic data for both language~\cite{gao2022self, honovich2023unnatural, meng2022generating, yang2022z, zhu2023visualize} and vision tasks~\cite{azizi2023synthetic, baradad2021learning, besnier2020dataset, he2022synthetic, hinterstoisser2018pre}. However, synthetic data can be noisy and inconsistently aligned, requiring noise-robust learning strategies~\cite{li2021learning, li2019learning, yao2020searching}, particularly in contrastive learning settings~\cite{liu2021noise, ibrahimi2022learning}. Prior work on synthetic compositional training highlights challenges in generating precise variations and maintaining cross-modal consistency.

Our approach differs in three key aspects. First, we generate multimodal video data from a single image with controllable object counts, spatial layouts, and temporal dynamics, enabling precise and structured supervision for multiple tasks. Second, the proposed pipeline supports unified training across diverse tasks, including object counting, video-based visual question answering, and video object segmentation. Third, we demonstrate strong transfer from predominantly synthetic training to real-world datasets, showing that carefully designed synthetic supervision can effectively bridge the synthetic-to-real gap.
%s\input{sec/3_backgound}
\begin{figure}[t]
    \centering
    \includegraphics[width=\linewidth]{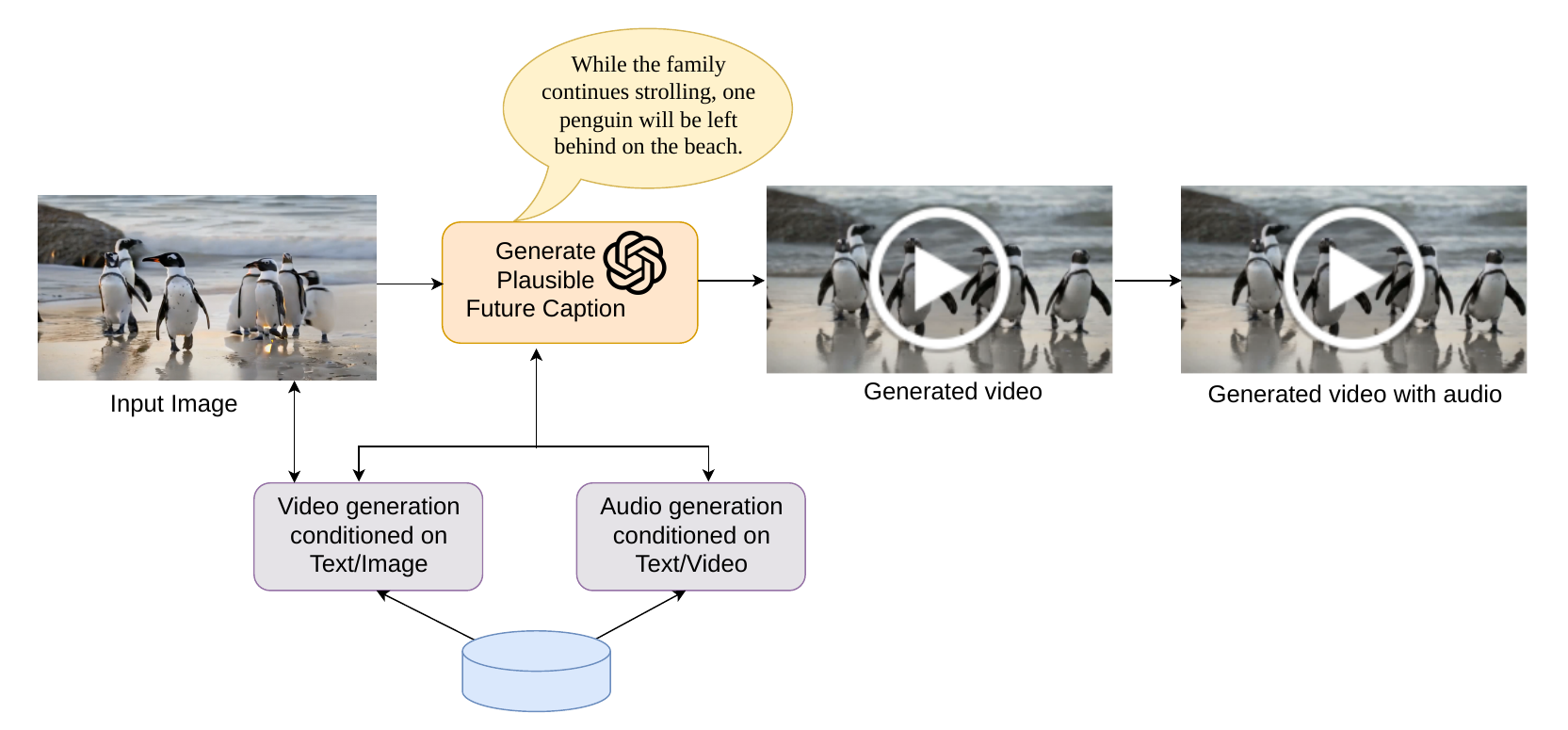}
    \caption{\textbf{Overview of the proposed multimodal synthetic data generation pipeline.} The pipeline takes a static Input Image from which an LLM generates a plausible future caption to establish temporal intent. The caption and image are used to condition a video generation model to produce a coherent visual sequence. Subsequently, an audio generation model, conditioned on both the text and the generated video, synthesizes a synchronized auditory track (optional). The final output is a high-fidelity generated video w/wo audio suitable for training MLLMs on complex temporal and cross-modal reasoning tasks.}
    \label{fig:pipeline}
    \vspace{-0.25in}
\end{figure}

\section{Synthetic Data Construction Pipeline}
\label{sec:methodology}

We propose a unified synthetic data generation pipeline that transforms a single (annotated)  input image -- data that is abundant -- into a set of semantically aligned multimodal outputs, including text, video, segmentation masks, and optionally audio. To reflect this design, we refer to the resulting synthetic dataset as \textbf{SynMulti} (Synthetic Multimodal Dataset). Unlike prior approaches that generate modalities independently, our framework models multimodal synthesis as a sequence of conditional generations, ensuring that each generated modality remains consistent with the others while being grounded in the original image.

Let \( I \in \mathcal{I} \) denote an input image. Our goal is to automatically expand one visual input into a structured multimodal sample:
\begin{equation}
\mathcal{D}(I) = \{ I, T, V, M, A \},
\end{equation}
where \( T \in \mathcal{T} \) is a textual description, \( V \in \mathcal{V} \) is a video sequence, \( M \in \mathcal{M} \) denotes segmentation masks, and \( A \in \mathcal{A} \) represents optional audio. We model the joint distribution as
\begin{equation}
p(T, V, M, A \mid I) = p(T \mid I)\, p(V \mid I, T)\, p(M \mid I, V)\, p(A \mid V, T).
\end{equation}
This factorization enforces sequential conditioning, allowing each modality to leverage previously generated outputs while maintaining alignment with the original image. In this way, SynMulti provides a controlled path from a single image to a richly annotated multimodal sample.

\vspace{-0.1in}
\subsection{Image as Primary Conditioning}
The input image \( I \) serves as the primary conditioning signal throughout the pipeline. It provides the visual grounding needed to preserve object identity, spatial layout, scene context, appearance attributes, and background structure. Beyond describing the observed scene, the image also serves as the basis for inferring plausible future events and temporal dynamics, enabling the generation of coherent dynamic content. Therefore, all generated modalities are conditioned on \( I \):
\begin{equation}
X \sim p(X \mid I, \cdot), \quad X \in \{T, V, M, A\}.
\end{equation}
Conditioning on the image throughout the generation process helps preserve object identity, scene geometry, and semantic consistency across modalities. It also reduces common generation artifacts such as identity drift, unrealistic object transformations, and implausible scene changes. For example, when a person appears in the input image, image-based conditioning encourages the generated video to maintain consistent visual characteristics for that individual over time.

\vspace{-0.1in}
\subsection{LLM-Guided Semantic Annotation}

\begin{figure}[t]
    \centering
    \includegraphics[width=\linewidth]{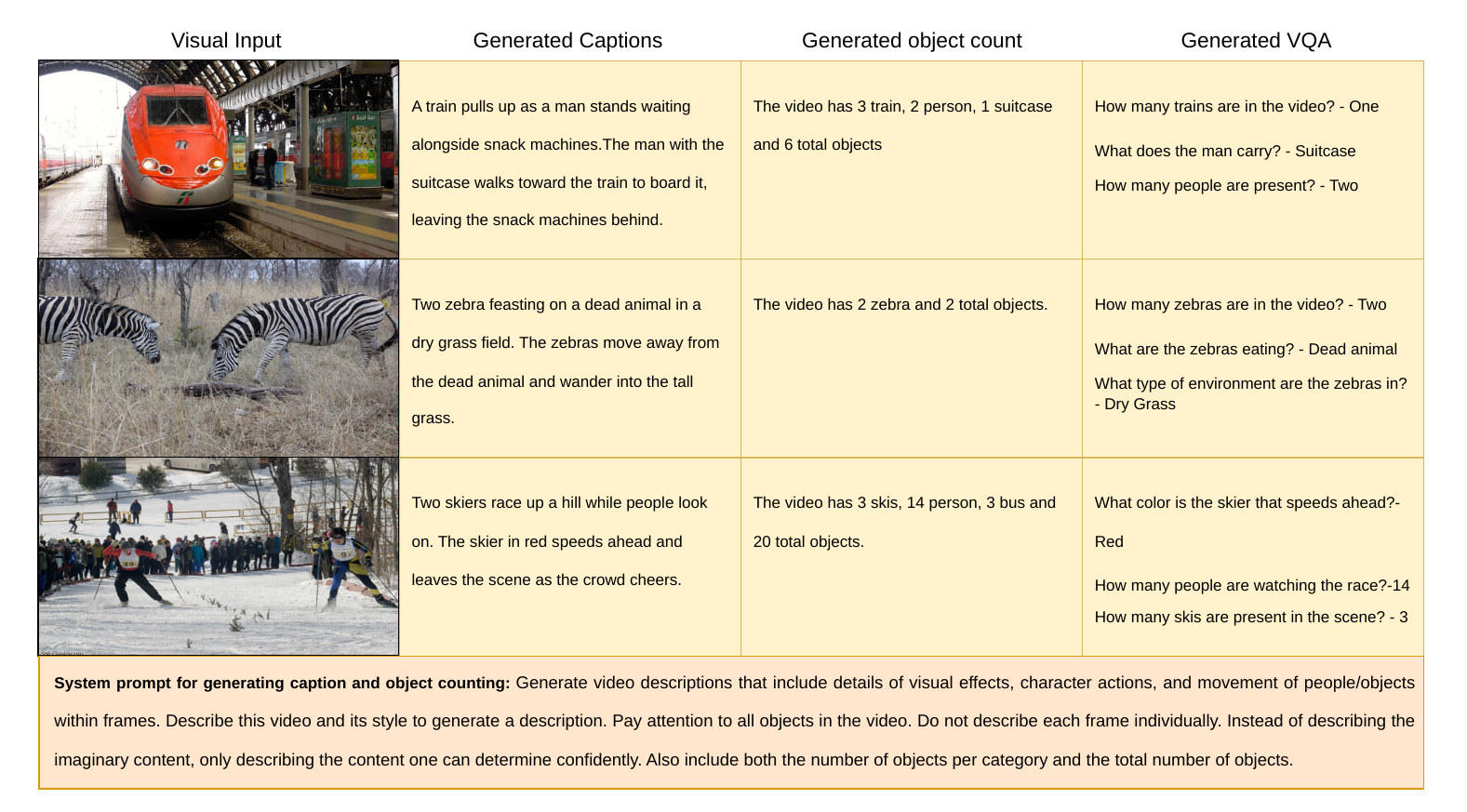}
    \caption{\textbf{Examples of multimodal annotations generated by our LLM-based pipeline. } Given a visual input, the system automatically produces (1) temporal captions describing the scene progression, (2) object-count annotations derived from detected entities, and (3) visual question–answer (VQA) pairs. This process enables scalable generation of diverse supervisory signals for vision-language learning.}
    \label{fig:text_annotation}
    \vspace{-0.3in}
\end{figure}

As illustrated in Figure~\ref{fig:text_annotation}, we begin by sampling images from the MSCOCO dataset~\cite{lin2014microsoft}, which provides diverse visual scenes together with object-level annotations. For each image \( I \), we construct a set of semantic annotations using a large language model (LLM), resulting in complementary supervision signals for captioning, counting, and visual reasoning. Therefore, first, as LLM we use ChatGPT~\cite{wu2023brief} to generate a semantic caption:
\begin{equation}
T = f_{\text{LLM}}(I),
\end{equation}
where \( T \) describes the visual content of the image, including objects, activities, scene context, and spatial relationships. In addition to observable information, the generated caption incorporates plausible future events that may naturally emerge from the scene. These future-aware captions provide temporal context that is later used to guide video generation from static images. The generated caption serves as a high-level semantic annotation that complements visual features by encoding contextual information, object interactions, and potential event progression. Such information helps bridge the gap between a static image and the dynamic content required for video synthesis. To further enrich supervision, we leverage the object annotations available in MSCOCO to derive object-count labels:
\begin{equation}
\mathcal{O}(I)=\{(c_k,n_k)\}_{k=1}^{K},
\end{equation}
where \( c_k \) denotes an object category and \( n_k \) represents the number of instances belonging to that category. These annotations provide explicit quantitative information regarding scene composition and object distributions. Specifically, we utilize instance-level annotations to compute the total number of objects and the number of instances for each object category within an image. This structured object-count information is then provided to LLM, which generates a natural language description summarizing the scene composition, such as the number of cars, the number of cats, and the total number of objects present in the image. We additionally generate visual question-answer (VQA) pairs using both the image and the generated caption:
\begin{equation}
\mathcal{Q}
=
\{(q_i,a_i)\}_{i=1}^{N}
=
f_{\text{VQA}}(I,T),
\end{equation}
where each question-answer pair requires reasoning about visual content, semantic context, or object relationships. By conditioning VQA generation on both generated caption and the image, we obtain richer reasoning-oriented supervision than would be possible from the image alone.

The complete textual supervision associated with each image is therefore defined as
\begin{equation}
\mathcal{S}_{\text{text}}
=
\{T,\mathcal{O}(I),\mathcal{Q}\}.
\end{equation}

Together, the generated captions, object-count annotations, and VQA pairs provide complementary semantic, quantitative, and reasoning-based signals. Examples of the generated annotations are shown in Figure~\ref{fig:text_annotation}. These annotations form the textual component of SynMulti and serve as supervision for training vision-language foundation models.

\vspace{-0.1in}
\subsection{Text- and Image-Conditioned Video Generation}

Given an input image \( I \) and its corresponding generated caption \( T \), we synthesize videos using a conditional video generation model that is conditioned on both modalities. Specifically, we adopt Wan2.2~\cite{wan2025wan}, which provides model variants with 5B and 14B parameters, and use the 14B variant in our experiments to achieve higher visual fidelity and improved temporal coherence. Formally, we generate a video \( V = \{v_t\}_{t=1}^{T_v} \) according to
\begin{equation}
V \sim p(V \mid I, T) = \prod_{t=1}^{T_v} p(v_t \mid v_{<t}, I, T),
\end{equation}
where \( v_{<t} \) denotes previously generated frames. In this formulation, the image \( I \) provides structural initialization and spatial grounding, while the caption \( T \) guides temporal dynamics and semantic evolution of the generated sequence. Conditioned on these inputs, the model generates videos that follow the implied scene dynamics and plausible future progression described by the caption, while remaining faithful to the visual structure of the initial frame. This design ensures both semantic consistency and realistic motion modeling. It also reduces drift in object appearance, camera layout, and scene composition over time.

As a result, we construct a multimodal dataset consisting of five components: (1) the generated video, (2) the generated corresponding semantic caption, (3) object-count annotations derived from the caption and underlying semantic content, and (4) visual question-answer (VQA) pairs.
Overall, this stage produces large-scale synthetic multimodal data with rich supervision for downstream vision-language and video understanding tasks, including visual question answering, object counting, and temporal reasoning.

\vspace{-0.1in}
\subsection{Audio as an Optional Modality}

In addition to visual modalities, our pipeline optionally generates synchronized audio to support multimodal data creation. Given a generated video \( V = \{v_t\}_{t=1}^{T_v} \) and its corresponding textual description \( T \), we employ VTA-LDM~\cite{xu2024vta-ldm} to synthesize an audio stream that is temporally aligned with the visual content. The generation process is formulated as
\begin{equation}
A = f_{\text{audio}}(V, T),
\end{equation}
corresponding to the conditional distribution
\begin{equation}
p(A \mid V, T),
\end{equation}
where \( A = \{a_t\}_{t=1}^{T_a} \) denotes the generated audio sequence.

This joint conditioning on visual dynamics and semantic captions ensures that the audio is both temporally synchronized with the video and semantically consistent with the scene description. For example, ocean wave scenes are paired with wave sounds, traffic scenes with vehicle noise and so on. By leveraging both motion cues from the video and semantic guidance from the text, the model generates contextually grounded audio rather than random synthesis. While our framework natively supports audio generation and can scale to comprehensive audiovisual datasets, we limit our current focus to visual tasks. We plan to extend our experiments to multimodal audiovisual tasks in future work.

\vspace{-0.1in}
\subsection{Mask Generation for Video Segmentation}

Accurate and temporally consistent segmentation masks are critical for advancing video understanding and generation. In addition to generating multiple modalities, we focus on mask generation as a key auxiliary modality for object-level segmentation, which provides high-quality synthetic data that can significantly enhance the training of video foundation models. By capturing both spatial and temporal object dynamics, these masks enable models to better understand motion, occlusion, and structural changes in videos, addressing one of the main challenges in video-centric learning. Starting from a single input image, we first generate precise, high-resolution object masks using SAM2~\cite{kirillov2023segment}. Let the input image be denoted by \( I \), and the initial set of object masks be represented as
\begin{equation}
M^{(0)} = \{ m_k^{(0)} \}_{k=1}^{K},
\end{equation}
where \( m_k^{(0)} \in \{0,1\}^{H \times W} \) denotes the binary mask of the \(k\)-th object in the initial frame. These masks are obtained by applying SAM2 as
\begin{equation}
M^{(0)} = f_{\text{SAM2}}(I).
\end{equation}
These masks are then propagated across generated video frames using MUG-VOS~\cite{lim2025multi}, ensuring temporal consistency and preserving object structures over time. Given a generated video \( V = \{ v_t \}_{t=1}^{T_v} \), we define the propagated mask sequence as
\begin{equation}
M^{(t)} = f_{\text{prop}}(v_t, v_{t-1}, M^{(t-1)}), \quad t=1,\ldots,T_v,
\end{equation}
where \( M^{(t)} = \{ m_k^{(t)} \}_{k=1}^{K} \) is the set of masks at frame \(t\). This transforms static masks into fully annotated video sequences that reflect object motion, appearance variation, and occlusion events. The resulting video segmentation masks can be utilized for a wide range of downstream applications. %They provide synthetic supervision for training video foundation models, enabling robust learning of spatio-temporal representations without manual annotation. They also improve video object tracking and action recognition by offering precise localization across frames, and facilitate video editing tasks such as selective object manipulation, background replacement, and scene-level compositing. Moreover, these masks support dataset augmentation and benchmarking, offering a scalable and automated approach to enrich video datasets while reducing dependence on costly human-labeled annotations.

\vspace{-0.1in}
\subsection{What Makes the Pipeline Unified}
We refer to our framework as a unified synthetic data pipeline because it formulates multimodal synthesis as a single image-anchored conditional process, rather than as a collection of independent generation tasks. This distinction is important: instead of producing text, video, masks, and audio through separate modules with weak coupling, SynMulti generates these outputs through a shared conditioning chain in which each modality is derived from the same visual source and the previously synthesized outputs. As a result, the entire pipeline is designed to preserve cross-modal alignment at the levels of object identity, scene structure, and temporal evolution.

Compared with conventional multimodal generators that synthesize modalities independently or with limited interaction between them, our framework supports flexible combinations such as image--text pairs, video--text pairs, video--mask annotations, and full multimodal vision--language--audio sequences within a single pipeline. This flexibility is valuable because it allows the same framework to serve multiple downstream training objectives while maintaining consistent supervision. In particular, the shared conditioning design provides fine-grained control over scene composition and temporal progression, which is difficult to achieve when modalities are generated in isolation.

The novelty of SynMulti lies in this unified formulation of synthetic data construction. By anchoring all generation to one input image and propagating semantic and structural information through successive modalities, the pipeline creates a coherent multimodal dataset rather than a set of loosely related outputs. This makes the generated data more suitable for training multimodal large language models, since the resulting samples provide aligned supervision across text, video, segmentation, and audio within a single coherent sample. The generated data can further facilitate synthetic-to-real transfer by exposing models to diverse, structured, and controllable visual reasoning scenarios that generalize to real-world video understanding tasks. In addition to improving dataset scalability, this unified construction also reduces modality fragmentation, since all annotations are derived from the same underlying scene representation. Therefore, SynMulti serves as a scalable and technically coherent solution for synthetic multimodal data generation.

\vspace{-0.1in}
\section{Experimental Results}
\label{sec:experiment}

\subsection{Training Using Synthetic Multimodal Data}

\begin{figure}[t]
    \centering
    \includegraphics[width=\linewidth]{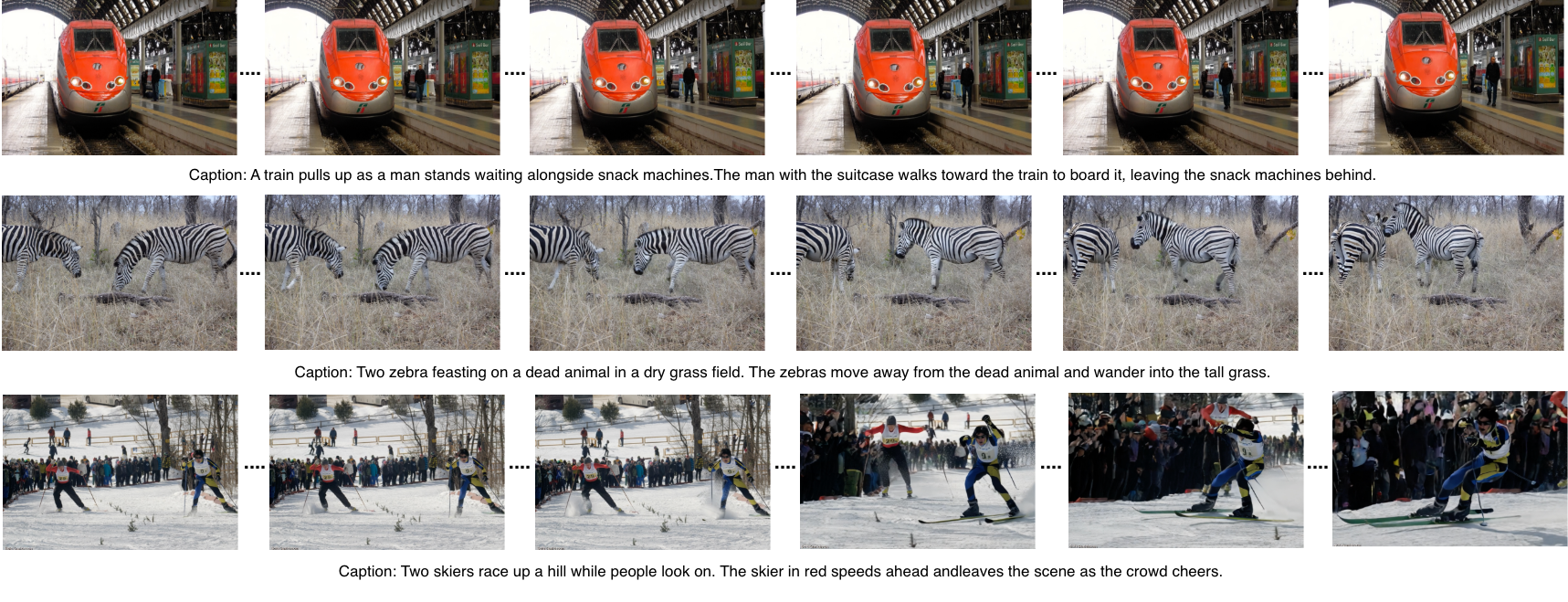}
    \caption{\textbf{Sample frames from our generated synthetic videos}, along with the corresponding captions produced by the LLM. These videos are automatically generated and annotated, providing temporally consistent frame-level information that is used to fine-tune the model. The visualization demonstrates that our synthetic pipeline produces diverse and semantically meaningful video content, which can serve as high-quality training data for enhancing the MLLM’s video understanding and captioning capabilities.}
    \label{fig:generated_videos}
    \vspace{-0.25in}
\end{figure}

Following the proposed synthetic data generation pipeline, we generate approximately 5K training videos and 1K validation videos using images from the MSCOCO dataset~\cite{lin2014microsoft}. These generated videos (see Figure~\ref{fig:generated_videos}), along with their associated multimodal annotations, form a synthetic dataset that is used to fine-tune a multimodal video foundation model.
In this work, we adopt InternVideo2.5~\cite{wang2025internvideo2} as the base video foundation model. To investigate the effectiveness of different supervision signals, we fine-tune the model using three types of training configurations: (1) caption-only supervision, following the conventional fine-tuning paradigm; (2) captions augmented with visual question–answer pairs; and (3) VQA-only supervision.
%Furthermore, we evaluate the fine-tuned model on two downstream tasks: object counting and visual question answering (VQA). For the counting task, the training annotations include both the total number of objects and the counts of individual object instances extracted from the captions.

The quantitative results for the counting and VQA tasks are presented in Table~\ref{tab:counting} and Table~\ref{tab:vqa}, respectively. In addition to the MSCOCO validation set, we also evaluate the model on the LV-VIS~\cite{wang2023towards} and YouTube-VIS~\cite{yang2019video} datasets to assess the generalization ability of models fine-tuned using synthetic data. 
Despite being trained solely on the synthetic dataset, the fine-tuned model consistently outperforms the baseline across multiple benchmarks, demonstrating the effectiveness of synthetic multimodal supervision for improving model performance.

\begin{table*}[t]
\centering
\scriptsize
\setlength{\tabcolsep}{8pt}
\begin{tabular}{llcc}
\toprule
\textbf{Dataset} & \textbf{Model} & \textbf{MAE} $\downarrow$ & \textbf{MSE} $\downarrow$ \\
\midrule

\multirow{4}{*}{MS-COCO}
& InternVideo2.5 & 5.46 & 75.38 \\
& + Fine-tuning (Captions) & 3.99 & 51.09 \\
& + Fine-tuning (Captions + VQA) & 3.30 & 29.50 \\
& + Fine-tuning (VQA) & \textbf{3.18} & \textbf{27.21} \\
\midrule

\multirow{4}{*}{LV-VIS}
& InternVideo2.5 & 0.97 & 36.18 \\
& + Fine-tuning (Captions) & 0.80 & 6.12 \\
& + Fine-tuning (Captions + VQA) & 0.79 & 5.22 \\
& + Fine-tuning (VQA) & \textbf{0.76} & \textbf{4.80} \\
\midrule

\multirow{4}{*}{YouTube-VIS}
& InternVideo2.5 & \textbf{0.39} & \textbf{0.71} \\
& + Fine-tuning (Captions) & 0.65 & 4.05 \\
& + Fine-tuning (Captions + VQA) & 0.75 & 4.87 \\
& + Fine-tuning (VQA) & 0.68 & 3.69 \\
\bottomrule
\end{tabular}
\vspace{0.1in}
\caption{\textbf{Video object counting results.} Mean Absolute Error (MAE) and Mean Squared Error (MSE) on the MS-COCO validation set, LV-VIS 2023, and YouTube-VIS datasets. Lower values indicate better counting accuracy. }
\vspace{-0.1in}
\label{tab:counting}
\end{table*}
\begin{table}[t]
\centering
\scriptsize
\setlength{\tabcolsep}{5pt}
\begin{tabular}{llcc}
\toprule
\textbf{Dataset} & \textbf{Model} & \textbf{CLIP-Score} $\uparrow$ & \textbf{WUP} $\uparrow$ \\
\midrule

\multirow{4}{*}{MS-COCO}
& InternVideo2.5 & 87.37 & 0.18 \\
& + Fine-tuning (Captions) & 85.32 & 0.20 \\
& + Fine-tuning (Captions + VQA) & \textbf{90.89} & \textbf{0.84} \\
& + Fine-tuning (VQA) & 90.23 & 0.81 \\
\midrule

\multirow{4}{*}{LV-VIS}
& InternVideo2.5 & 75.16 & 0.45 \\
& + Fine-tuning (Captions) & 69.54 & 0.25 \\
& + Fine-tuning (Captions + VQA) & \textbf{77.91} & 0.54 \\
& + Fine-tuning (VQA) & 76.51 & \textbf{0.58} \\
\midrule

\multirow{4}{*}{YouTube-VIS}
& InternVideo2.5 & 82.37 & 0.40 \\
& + Fine-tuning (Captions) & 77.05 & 0.20 \\
& + Fine-tuning (Captions + VQA) & \textbf{83.12} & 0.46 \\
& + Fine-tuning (VQA) & 80.85 & \textbf{0.48} \\
\bottomrule
\end{tabular}
\vspace{0.1in}
\caption{\textbf{Video question answering results.} CLIP-Score and Wu-Palmer Similarity (WUP) on MS-COCO, LV-VIS, and YouTube-VIS. Higher values indicate better semantic alignment and answer correctness. }
\label{tab:vqa}
\vspace{-0.35in}
\end{table}

\begin{table}[h]
\centering
\scriptsize
\begin{tabular}{lccc}
\toprule
\textbf{Class} & \textbf{Baseline} & \textbf{Ours} & $\Delta$ \\
\midrule
Toilet        & 0.10 & 0.79 & +0.69 \\
Sink          & 0.13 & 0.87 & +0.74 \\
Bed           & 0.06 & 0.99 & +0.93 \\
Bicycle       & 0.26 & 0.83 & +0.57 \\
Car           & 0.09 & 0.56 & +0.47 \\
Dog           & 0.46 & 0.75 & +0.29 \\
\midrule
Microwave     & 0.81 & 0.68 & -0.13 \\
Apple         & 0.80 & 0.74 & -0.06 \\
Cake          & 0.91 & 0.55 & -0.36 \\
Knife         & 0.69 & 0.60 & -0.09 \\
Train         & 0.73 & 0.45 & -0.28 \\
Surfboard     & 0.69 & 0.28 & -0.41 \\
\midrule
\textbf{mIoU} & \textbf{0.4711} & \textbf{0.5239} & \textbf{+0.0528} \\
\midrule
\multicolumn{4}{l}{\textbf{Improved: 36 \quad Degraded: 26 \quad Unchanged: 12}} \\
\bottomrule
\end{tabular}
\vspace{0.1in}
\caption{Per-class IoU comparison on the MS COCO dataset. We show representative improved and moderately degraded classes (excluding extreme failures where IoU drops to zero for clarity). The mean IoU (mIoU) is computed over all 74 classes. Synthetic fine-tuning improves overall performance by +5.28 mIoU.}
\label{tab:segmentation_baseline}
\vspace{-0.4in}
\end{table}

\vspace{-0.1in}
\subsection{Evaluation Metrics}
We evaluate our models using task-specific metrics to comprehensively measure performance. For video object counting (in Table~\ref{tab:counting}), we report the mean absolute error (MAE) and the mean squared error (MSE) ~\cite{hodson2022root}, which quantify the deviation between the predicted and ground-truth object counts, where lower values indicate better performance. Fine-tuning on synthetic temporal supervision substantially improves counting performance on MS-COCO and LV-VIS, while the pretrained InternVideo2.5 model achieves the best performance on YouTube-VIS. For video-VQA (in Table~\ref{tab:vqa}), we adopt CLIP-Score~\cite{kang2025clip} and Word Understanding Precision (WUP)~\cite{choi2014text} to assess the semantic alignment and correctness of the generated answers with respect to the video content, with higher values indicating better performance. Same as object counting, fine-tuning with synthetic VQA supervision consistently improves performance, while combining caption and VQA supervision yields the strongest overall CLIP-Score. For segmentation (in Table~\ref{tab:segmentation_baseline} and ~\ref{tab:segmentation_scale}), we use mean Intersection over Union (mIoU) to evaluate the per-class and overall segmentation accuracy, where higher values reflect better overlap between the predicted and ground-truth masks. These metrics provide a comprehensive assessment of our framework for counting, reasoning, and visual understanding tasks.

\vspace{-0.1in}
\subsection{Effect of Synthetic Data on Segmentation}

We evaluate the effectiveness of synthetic video data for improving segmentation performance. Synthetic training videos are generated using our pipeline, which provides temporally consistent object-level masks across frames, and are used to fine-tune SAM.

Fine-tuning with synthetic videos leads to consistent improvements over the baseline. We further analyze the impact of data scale by training with 2K and 5K videos. As shown in Tables~\ref{tab:segmentation_baseline} and~\ref{tab:segmentation_scale}, increasing the amount of synthetic data improves performance, with the 5K setting achieving higher mean IoU than the 2K setting. Some classes exhibit minor degradations after fine-tuning, likely due to domain shift, overfitting to synthetic patterns, or mask propagation errors. These effects can be mitigated by incorporating a small amount of real annotated data, increasing data diversity, or improving mask quality. Overall, the results demonstrate consistent gains in mean IoU.

These findings highlight the scalability of our approach, where large-scale synthetic data improves segmentation performance while reducing the reliance on manual annotation.

\begin{table}[h]
\centering
\scriptsize
\begin{tabular}{lccc}
\toprule
\textbf{Class} & \textbf{2K Videos} & \textbf{5K Videos} & $\Delta$ (5K - 2K) \\
\midrule
Toilet        & 0.10 & 0.79 & +0.69 \\
Sink          & 0.15 & 0.87 & +0.72 \\
Bed           & 0.50 & 0.99 & +0.49 \\
Bicycle       & 0.65 & 0.83 & +0.18 \\
Car           & 0.20 & 0.56 & +0.36 \\
Dog           & 0.48 & 0.75 & +0.27 \\
\midrule
Microwave     & 0.84 & 0.68 & -0.16 \\
Cake          & 0.77 & 0.55 & -0.22 \\
Knife         & 0.62 & 0.60 & -0.02 \\
Train         & 0.74 & 0.45 & -0.29 \\
Surfboard     & 0.58 & 0.28 & -0.30 \\
\midrule
\textbf{mIoU} & \textbf{0.4694} & \textbf{0.5239} & \textbf{+0.0545} \\
\bottomrule
\end{tabular}
\vspace{0.1in}
\caption{Effect of synthetic data scale on segmentation performance. Increasing training data from 2K to 5K videos improves mIoU from 0.4694 to 0.5239 (+5.45). Results are reported on MSCOCO (74 classes), with representative classes shown for clarity.}
\label{tab:segmentation_scale}
\vspace{-0.3in}
\end{table}

\vspace{-0.1in}
\subsection{Robust Learning from Imperfect Synthetic Video Data} 
Synthetic videos generated by our pipeline are not flawless; they may exhibit minor artifacts, unrealistic object motions, or occasional misalignments. Despite these limitations, they provide valuable supervision for model training. The primary advantage lies in their diversity: synthetic data introduces a wide range of objects, scenes, and actions that are often underrepresented in real-world datasets, enabling the model to learn richer visual–language correspondences. Moreover, temporally consistent annotations propagated across frames offer dense supervision that is rarely available in large-scale real video corpora. This consistency helps the model learn stable object representations over time. In addition, the scale of synthetic data exposes the model to thousands of variations, improving generalization even when individual samples are imperfect. 

When used for fine-tuning, synthetic videos complement knowledge acquired from real datasets, leading to improved performance on downstream tasks such as video captioning, visual question answering, object counting, and segmentation. Overall, imperfect synthetic videos serve as a cost-effective and scalable source of training data that reliably enhances model performance.

\section{Conclusions}
\label{sec:conclusion}
In this work, we propose the SynMulti dataset, along with a scalable pipeline to generate synthetic videos, and systematically study how such data can be leveraged to train multimodal vision–language models. By producing temporally consistent videos with frame-level annotations, SynthMulti provides a cost-effective alternative to manual data collection. Through extensive experiments on video VQA, object counting, and segmentation, we demonstrate that fine-tuning with synthetic data consistently improves performance across tasks. Furthermore, we show that increasing the volume of synthetic data yields additional performance gains, highlighting the scalability of our framework. Our analysis further reveals that even imperfect synthetic videos offer valuable supervision. They introduce diversity and temporal consistency that are often lacking in real-world datasets, enabling models to learn richer and more stable representations. These findings position synthetic data not only as a supplement but as a key component for training large-scale multimodal models. In future work, we plan to extend SynthMulti to audio-visual settings by incorporating synthetic audio, enabling more comprehensive multimodal reasoning.

% ---- Bibliography ----
%
% BibTeX users should specify bibliography style 'splncs04'.
% References will then be sorted and formatted in the correct style.
%
\bibliographystyle{splncs04}
\bibliography{main}

@String(ECCV  = {Eur. Conf. Comput. Vis.})

@String(NeurIPS = {Adv. Neural Inform. Process. Syst.})

@String(AAAI  = {AAAI})

@String(ICASSP=	{ICASSP})

@String(ECCV  = {ECCV})

@String(NeurIPS = {NeurIPS})

@article{nguyen2023improving,
  title={Improving multimodal datasets with image captioning},
  author={Nguyen, Thao and Gadre, Samir Yitzhak and Ilharco, Gabriel and Oh, Sewoong and Schmidt, Ludwig},
  journal={NeurIPS},
  volume={36},
  pages={22047--22069},
  year={2023}
}

@inproceedings{hu2023promptcap,
  title={Promptcap: Prompt-guided image captioning for vqa with gpt-3},
  author={Hu, Yushi and Hua, Hang and Yang, Zhengyuan and Shi, Weijia and Smith, Noah A and Luo, Jiebo},
  booktitle={Proceedings of the IEEE/CVF International Conference on Computer Vision},
  pages={2963--2975},
  year={2023}
}

@inproceedings{jia2025vqa2,
  title={Vqa2: visual question answering for video quality assessment},
  author={Jia, Ziheng and Zhang, Zicheng and Qian, Jiaying and Wu, Haoning and Sun, Wei and Li, Chunyi and Liu, Xiaohong and Lin, Weisi and Zhai, Guangtao and Min, Xiongkuo},
  booktitle={Proceedings of the 33rd ACM International Conference on Multimedia},
  pages={6751--6760},
  year={2025}
}

@article{tiong2022plug,
  title={Plug-and-play vqa: Zero-shot vqa by conjoining large pretrained models with zero training},
  author={Tiong, Anthony Meng Huat and Li, Junnan and Li, Boyang and Savarese, Silvio and Hoi, Steven CH},
  journal={arXiv preprint arXiv:2210.08773},
  year={2022}
}

@inproceedings{changpinyo2022all,
  title={All you may need for VQA are image captions},
  author={Changpinyo, Soravit and Kukliansy, Doron and Szpektor, Idan and Chen, Xi and Ding, Nan and Soricut, Radu},
  booktitle={Proceedings of the 2022 conference of the north american chapter of the association for computational linguistics: human language technologies},
  pages={1947--1963},
  year={2022}
}

@inproceedings{hu2022scaling,
  title={Scaling up vision-language pre-training for image captioning},
  author={Hu, Xiaowei and Gan, Zhe and Wang, Jianfeng and Yang, Zhengyuan and Liu, Zicheng and Lu, Yumao and Wang, Lijuan},
  booktitle={Proceedings of the IEEE/CVF conference on computer vision and pattern recognition},
  pages={17980--17989},
  year={2022}
}

@article{li2025videochat,
  title={Videochat: Chat-centric video understanding},
  author={Li, KunChang and He, Yinan and Wang, Yi and Li, Yizhuo and Wang, Wenhai and Luo, Ping and Wang, Yali and Wang, Limin and Qiao, Yu},
  journal={Science China Information Sciences},
  volume={68},
  number={10},
  pages={200102},
  year={2025},
  publisher={Springer}
}

@inproceedings{buch2022revisiting,
  title={Revisiting the" video" in video-language understanding},
  author={Buch, Shyamal and Eyzaguirre, Crist{\'o}bal and Gaidon, Adrien and Wu, Jiajun and Fei-Fei, Li and Niebles, Juan Carlos},
  booktitle={Proceedings of the IEEE/CVF conference on computer vision and pattern recognition},
  pages={2917--2927},
  year={2022}
}

@article{tang2025video,
  title={Video understanding with large language models: A survey},
  author={Tang, Yunlong and Bi, Jing and Xu, Siting and Song, Luchuan and Liang, Susan and Wang, Teng and Zhang, Daoan and An, Jie and Lin, Jingyang and Zhu, Rongyi and others},
  journal={IEEE Transactions on Circuits and Systems for Video Technology},
  year={2025},
  publisher={IEEE}
}

@inproceedings{lin2014microsoft,
  title={Microsoft coco: Common objects in context},
  author={Lin, Tsung-Yi and Maire, Michael and Belongie, Serge and Hays, James and Perona, Pietro and Ramanan, Deva and Doll{\'a}r, Piotr and Zitnick, C Lawrence},
  booktitle={European conference on computer vision},
  pages={740--755},
  year={2014},
  organization={Springer}
}

@article{krishna2017visual,
  title={Visual genome: Connecting language and vision using crowdsourced dense image annotations},
  author={Krishna, Ranjay and Zhu, Yuke and Groth, Oliver and Johnson, Justin and Hata, Kenji and Kravitz, Joshua and Chen, Stephanie and Kalantidis, Yannis and Li, Li-Jia and Shamma, David A and others},
  journal={International journal of computer vision},
  volume={123},
  number={1},
  pages={32--73},
  year={2017},
  publisher={Springer}
}

@inproceedings{li2023uniformerv2,
  title={Uniformerv2: Unlocking the potential of image vits for video understanding},
  author={Li, Kunchang and Wang, Yali and He, Yinan and Li, Yizhuo and Wang, Yi and Wang, Limin and Qiao, Yu},
  booktitle={Proceedings of the IEEE/CVF International Conference on Computer Vision},
  pages={1632--1643},
  year={2023}
}

@article{soomro2012ucf101,
  title={Ucf101: A dataset of 101 human actions classes from videos in the wild},
  author={Soomro, Khurram and Zamir, Amir Roshan and Shah, Mubarak},
  journal={arXiv preprint arXiv:1212.0402},
  year={2012}
}

@inproceedings{xu2018youtube,
  title={Youtube-vos: Sequence-to-sequence video object segmentation},
  author={Xu, Ning and Yang, Linjie and Fan, Yuchen and Yang, Jianchao and Yue, Dingcheng and Liang, Yuchen and Price, Brian and Cohen, Scott and Huang, Thomas},
  booktitle={Proceedings of the European conference on computer vision (ECCV)},
  pages={585--601},
  year={2018}
}

@article{damen2020epic,
  title={The epic-kitchens dataset: Collection, challenges and baselines},
  author={Damen, Dima and Doughty, Hazel and Farinella, Giovanni Maria and Fidler, Sanja and Furnari, Antonino and Kazakos, Evangelos and Moltisanti, Davide and Munro, Jonathan and Perrett, Toby and Price, Will and others},
  journal={IEEE Transactions on Pattern Analysis and Machine Intelligence},
  volume={43},
  number={11},
  pages={4125--4141},
  year={2020},
  publisher={IEEE}
}

@inproceedings{pothiraj2025capture,
  title={Capture: Evaluating spatial reasoning in vision language models via occluded object counting},
  author={Pothiraj, Atin and Stengel-Eskin, Elias and Cho, Jaemin and Bansal, Mohit},
  booktitle={Proceedings of the IEEE/CVF International Conference on Computer Vision},
  pages={8001--8010},
  year={2025}
}

@inproceedings{wang2024language,
  title={Language-guided zero-shot object counting},
  author={Wang, Mingjie and Yuan, Song and Li, Zhuohang and Zhu, Longlong and Buys, Eric and Gong, Minglun},
  booktitle={2024 IEEE International Conference on Multimedia and Expo Workshops (ICMEW)},
  pages={1--6},
  year={2024},
  organization={IEEE}
}

@inproceedings{binyamin2025make,
  title={Make it count: Text-to-image generation with an accurate number of objects},
  author={Binyamin, Lital and Tewel, Yoad and Segev, Hilit and Hirsch, Eran and Rassin, Royi and Chechik, Gal},
  booktitle={Proceedings of the Computer Vision and Pattern Recognition Conference},
  pages={13242--13251},
  year={2025}
}

@inproceedings{antol2015vqa,
  title={Vqa: Visual question answering},
  author={Antol, Stanislaw and Agrawal, Aishwarya and Lu, Jiasen and Mitchell, Margaret and Batra, Dhruv and Zitnick, C Lawrence and Parikh, Devi},
  booktitle={Proceedings of the IEEE international conference on computer vision},
  pages={2425--2433},
  year={2015}
}

@inproceedings{goyal2017making,
  title={Making the v in vqa matter: Elevating the role of image understanding in visual question answering},
  author={Goyal, Yash and Khot, Tejas and Summers-Stay, Douglas and Batra, Dhruv and Parikh, Devi},
  booktitle={Proceedings of the IEEE conference on computer vision and pattern recognition},
  pages={6904--6913},
  year={2017}
}

@inproceedings{singh2019towards,
  title={Towards vqa models that can read},
  author={Singh, Amanpreet and Natarajan, Vivek and Shah, Meet and Jiang, Yu and Chen, Xinlei and Batra, Dhruv and Parikh, Devi and Rohrbach, Marcus},
  booktitle={Proceedings of the IEEE/CVF conference on computer vision and pattern recognition},
  pages={8317--8326},
  year={2019}
}

@article{yao2020video,
  title={Video object segmentation and tracking: A survey},
  author={Yao, Rui and Lin, Guosheng and Xia, Shixiong and Zhao, Jiaqi and Zhou, Yong},
  journal={ACM Transactions on Intelligent Systems and Technology (TIST)},
  volume={11},
  number={4},
  pages={1--47},
  year={2020},
  publisher={ACM New York, NY, USA}
}

@article{gao2023deep,
  title={Deep learning for video object segmentation: a review},
  author={Gao, Mingqi and Zheng, Feng and Yu, James JQ and Shan, Caifeng and Ding, Guiguang and Han, Jungong},
  journal={Artificial Intelligence Review},
  volume={56},
  number={1},
  pages={457--531},
  year={2023},
  publisher={Springer}
}

@inproceedings{caelles2017one,
  title={One-shot video object segmentation},
  author={Caelles, Sergi and Maninis, Kevis-Kokitsi and Pont-Tuset, Jordi and Leal-Taix{\'e}, Laura and Cremers, Daniel and Van Gool, Luc},
  booktitle={Proceedings of the IEEE conference on computer vision and pattern recognition},
  pages={221--230},
  year={2017}
}

@article{wang2025clipcap,
  title={ClipCap++: An efficient image captioning approach via image encoder optimization and LLM fine-tuning},
  author={Wang, Ruiqin and Wu, Ye and Sheng, Zhenzhen},
  journal={Applied Soft Computing},
  volume={180},
  pages={113469},
  year={2025},
  publisher={Elsevier}
}

@article{somepalli2024calvin,
  title={CALVIN: Improved contextual video captioning via instruction tuning},
  author={Somepalli, Gowthami and Chowdhury, Arkabandhu and Geiping, Jonas and Basri, Ronen and Goldstein, Tom and Jacobs, David},
  journal={Advances in Neural Information Processing Systems},
  volume={37},
  pages={92983--93010},
  year={2024}
}

@inproceedings{zhang2024qwen,
  title={Qwen-IG: a Qwen-based instruction generation model for LLM fine-tuning},
  author={Zhang, Lu and Liu, Yu and Luo, Yitian and Gao, Feng and Gu, Jinguang},
  booktitle={Proceedings of the 2024 13th International Conference on Computing and Pattern Recognition},
  pages={295--302},
  year={2024}
}

@article{chawla2002smote,
  title={SMOTE: synthetic minority over-sampling technique},
  author={Chawla, Nitesh V and Bowyer, Kevin W and Hall, Lawrence O and Kegelmeyer, W Philip},
  journal={Journal of artificial intelligence research},
  volume={16},
  pages={321--357},
  year={2002}
}

@inproceedings{liu2023revisiting,
  title={Revisiting temporal modeling for clip-based image-to-video knowledge transferring},
  author={Liu, Ruyang and Huang, Jingjia and Li, Ge and Feng, Jiashi and Wu, Xinglong and Li, Thomas H},
  booktitle={Proceedings of the IEEE/CVF Conference on Computer Vision and Pattern Recognition},
  pages={6555--6564},
  year={2023}
}

@article{isobe2020revisiting,
  title={Revisiting temporal modeling for video super-resolution},
  author={Isobe, Takashi and Zhu, Fang and Jia, Xu and Wang, Shengjin},
  journal={arXiv preprint arXiv:2008.05765},
  year={2020}
}

@article{ravishankara2025artificial,
  title={The Artificial Intelligence Cognitive Examination: A Survey on the Evolution of Multimodal Evaluation From Recognition to Reasoning},
  author={Ravishankara, Mayank and Maharaj, Varindra V Persad},
  journal={IEEE Access},
  volume={14},
  pages={2690--2725},
  year={2025},
  publisher={IEEE}
}

@article{wang2024exploring,
  title={Exploring the reasoning abilities of multimodal large language models (mllms): A comprehensive survey on emerging trends in multimodal reasoning},
  author={Wang, Yiqi and Chen, Wentao and Han, Xiaotian and Lin, Xudong and Zhao, Haiteng and Liu, Yongfei and Zhai, Bohan and Yuan, Jianbo and You, Quanzeng and Yang, Hongxia},
  journal={arXiv preprint arXiv:2401.06805},
  year={2024}
}

@article{de2020overview,
  title={An overview of privacy in machine learning},
  author={De Cristofaro, Emiliano},
  journal={arXiv preprint arXiv:2005.08679},
  year={2020}
}

@article{xu2021privacy,
  title={Privacy-preserving machine learning: Methods, challenges and directions},
  author={Xu, Runhua and Baracaldo, Nathalie and Joshi, James},
  journal={arXiv preprint arXiv:2108.04417},
  year={2021}
}

@article{diederik2019introduction,
  title={An introduction to variational autoencoders},
  author={Diederik, P Kingma and Max, Welling},
  journal={Foundations and Trends{\textregistered} in Machine Learning},
  volume={12},
  number={4},
  pages={307--392},
  year={2019},
  publisher={Emerald Publishing Limited}
}

@article{goodfellow2020generative,
  title={Generative adversarial networks},
  author={Goodfellow, Ian and Pouget-Abadie, Jean and Mirza, Mehdi and Xu, Bing and Warde-Farley, David and Ozair, Sherjil and Courville, Aaron and Bengio, Yoshua},
  journal={Communications of the ACM},
  volume={63},
  number={11},
  pages={139--144},
  year={2020},
  publisher={ACM New York, NY, USA}
}

@article{gragnaniello2021gan,
  title={Are GAN generated images easy to detect? A critical analysis of the state-of-the-art},
  author={Gragnaniello, Diego and Cozzolino, Davide and Marra, Francesco and Poggi, Giovanni and Verdoliva, Luisa},
  journal={arXiv preprint arXiv:2104.02617},
  year={2021}
}

@inproceedings{hou2023evading,
  title={Evading deepfake detectors via adversarial statistical consistency},
  author={Hou, Yang and Guo, Qing and Huang, Yihao and Xie, Xiaofei and Ma, Lei and Zhao, Jianjun},
  booktitle={Proceedings of the IEEE/CVF conference on computer vision and pattern recognition},
  pages={12271--12280},
  year={2023}
}

@inproceedings{barni2020cnn,
  title={CNN detection of GAN-generated face images based on cross-band co-occurrences analysis},
  author={Barni, Mauro and Kallas, Kassem and Nowroozi, Ehsan and Tondi, Benedetta},
  booktitle={2020 IEEE international workshop on information forensics and security (WIFS)},
  pages={1--6},
  year={2020},
  organization={IEEE}
}

@inproceedings{frank2020leveraging,
  title={Leveraging frequency analysis for deep fake image recognition},
  author={Frank, Joel and Eisenhofer, Thorsten and Sch{\"o}nherr, Lea and Fischer, Asja and Kolossa, Dorothea and Holz, Thorsten},
  booktitle={International conference on machine learning},
  pages={3247--3258},
  year={2020},
  organization={PMLR}
}

@article{ji2024distinguish,
  title={Distinguish any fake videos: Unleashing the power of large-scale data and motion features},
  author={Ji, Lichuan and Lin, Yingqi and Huang, Zhenhua and Han, Yan and Xu, Xiaogang and Wu, Jiafei and Wang, Chong and Liu, Zhe},
  journal={arXiv preprint arXiv:2405.15343},
  year={2024}
}

@inproceedings{bai2024ai,
  title={Ai-generated video detection via spatial-temporal anomaly learning},
  author={Bai, Jianfa and Lin, Man and Cao, Gang and Lou, Zijie},
  booktitle={Chinese Conference on Pattern Recognition and Computer Vision (PRCV)},
  pages={460--470},
  year={2024},
  organization={Springer}
}

@inproceedings{dong2022explaining,
  title={Explaining deepfake detection by analysing image matching},
  author={Dong, Shichao and Wang, Jin and Liang, Jiajun and Fan, Haoqiang and Ji, Renhe},
  booktitle={European conference on computer vision},
  pages={18--35},
  year={2022},
  organization={Springer}
}

@inproceedings{chai2020makes,
  title={What makes fake images detectable? understanding properties that generalize},
  author={Chai, Lucy and Bau, David and Lim, Ser-Nam and Isola, Phillip},
  booktitle={European conference on computer vision},
  pages={103--120},
  year={2020},
  organization={Springer}
}

@inproceedings{shao2023detecting,
  title={Detecting and grounding multi-modal media manipulation},
  author={Shao, Rui and Wu, Tianxing and Liu, Ziwei},
  booktitle={Proceedings of the IEEE/CVF Conference on Computer Vision and Pattern Recognition},
  pages={6904--6913},
  year={2023}
}

@article{lu2023seeing,
  title={Seeing is not always believing: Benchmarking human and model perception of ai-generated images},
  author={Lu, Zeyu and Huang, Di and Bai, Lei and Qu, Jingjing and Wu, Chengyue and Liu, Xihui and Ouyang, Wanli},
  journal={Advances in neural information processing systems},
  volume={36},
  pages={25435--25447},
  year={2023}
}

@article{guo2023close,
  title={How close is chatgpt to human experts? comparison corpus, evaluation, and detection},
  author={Guo, Biyang and Zhang, Xin and Wang, Ziyuan and Jiang, Minqi and Nie, Jinran and Ding, Yuxuan and Yue, Jianwei and Wu, Yupeng},
  journal={arXiv preprint arXiv:2301.07597},
  year={2023}
}

@article{wang2020asvspoof,
  title={ASVspoof 2019: A large-scale public database of synthesized, converted and replayed speech},
  author={Wang, Xin and Yamagishi, Junichi and Todisco, Massimiliano and Delgado, H{\'e}ctor and Nautsch, Andreas and Evans, Nicholas and Sahidullah, Md and Vestman, Ville and Kinnunen, Tomi and Lee, Kong Aik and others},
  journal={Computer Speech \& Language},
  volume={64},
  pages={101114},
  year={2020},
  publisher={Elsevier}
}

@inproceedings{gani2025vane,
  title={Vane-bench: Video anomaly evaluation benchmark for conversational lmms},
  author={Gani, Hanan and Bharadwaj, Rohit and Naseer, Muzammal and Khan, Fahad Shahbaz and Khan, Salman},
  booktitle={Findings of the Association for Computational Linguistics: NAACL 2025},
  pages={3123--3140},
  year={2025}
}

@article{li2025fakebench,
  title={Fakebench: Probing explainable fake image detection via large multimodal models},
  author={Li, Yixuan and Liu, Xuelin and Wang, Xiaoyang and Lee, Bu Sung and Wang, Shiqi and Rocha, Anderson and Lin, Weisi},
  journal={IEEE Transactions on Information Forensics and Security},
  year={2025},
  publisher={IEEE}
}

@article{ye2024loki,
  title={Loki: A comprehensive synthetic data detection benchmark using large multimodal models},
  author={Ye, Junyan and Zhou, Baichuan and Huang, Zilong and Zhang, Junan and Bai, Tianyi and Kang, Hengrui and He, Jun and Lin, Honglin and Wang, Zihao and Wu, Tong and others},
  journal={arXiv preprint arXiv:2410.09732},
  year={2024}
}

@article{lu2022learn,
  title={Learn to explain: Multimodal reasoning via thought chains for science question answering},
  author={Lu, Pan and Mishra, Swaroop and Xia, Tanglin and Qiu, Liang and Chang, Kai-Wei and Zhu, Song-Chun and Tafjord, Oyvind and Clark, Peter and Kalyan, Ashwin},
  journal={Advances in neural information processing systems},
  volume={35},
  pages={2507--2521},
  year={2022}
}

@inproceedings{yue2024mmmu,
  title={Mmmu: A massive multi-discipline multimodal understanding and reasoning benchmark for expert agi},
  author={Yue, Xiang and Ni, Yuansheng and Zhang, Kai and Zheng, Tianyu and Liu, Ruoqi and Zhang, Ge and Stevens, Samuel and Jiang, Dongfu and Ren, Weiming and Sun, Yuxuan and others},
  booktitle={Proceedings of the IEEE/CVF conference on computer vision and pattern recognition},
  pages={9556--9567},
  year={2024}
}

@article{bai2024survey,
  title={A survey of multimodal large language model from a data-centric perspective},
  author={Bai, Tianyi and Liang, Hao and Wan, Binwang and Xu, Yanran and Li, Xi and Li, Shiyu and Yang, Ling and Li, Bozhou and Wang, Yifan and Cui, Bin and others},
  journal={arXiv preprint arXiv:2405.16640},
  year={2024}
}

@inproceedings{kamath2023s,
  title={What’s “up” with vision-language models? investigating their struggle with spatial reasoning},
  author={Kamath, Amita and Hessel, Jack and Chang, Kai-Wei},
  booktitle={Proceedings of the 2023 Conference on Empirical Methods in Natural Language Processing},
  pages={9161--9175},
  year={2023}
}

@inproceedings{peng2024synthesize,
  title={Synthesize diagnose and optimize: Towards fine-grained vision-language understanding},
  author={Peng, Wujian and Xie, Sicheng and You, Zuyao and Lan, Shiyi and Wu, Zuxuan},
  booktitle={Proceedings of the IEEE/CVF Conference on Computer Vision and Pattern Recognition},
  pages={13279--13288},
  year={2024}
}

@article{yuksekgonul2022and,
  title={When and why vision-language models behave like bags-of-words, and what to do about it?},
  author={Yuksekgonul, Mert and Bianchi, Federico and Kalluri, Pratyusha and Jurafsky, Dan and Zou, James},
  journal={arXiv preprint arXiv:2210.01936},
  year={2022}
}

@inproceedings{thrush2022winoground,
  title={Winoground: Probing vision and language models for visio-linguistic compositionality},
  author={Thrush, Tristan and Jiang, Ryan and Bartolo, Max and Singh, Amanpreet and Williams, Adina and Kiela, Douwe and Ross, Candace},
  booktitle={Proceedings of the IEEE/CVF Conference on Computer Vision and Pattern Recognition},
  pages={5238--5248},
  year={2022}
}

@inproceedings{wang2023self,
  title={Self-instruct: Aligning language models with self-generated instructions},
  author={Wang, Yizhong and Kordi, Yeganeh and Mishra, Swaroop and Liu, Alisa and Smith, Noah A and Khashabi, Daniel and Hajishirzi, Hannaneh},
  booktitle={Proceedings of the 61st annual meeting of the association for computational linguistics (volume 1: long papers)},
  pages={13484--13508},
  year={2023}
}

@article{zhao2022vl,
  title={Vl-checklist: Evaluating pre-trained vision-language models with objects, attributes and relations},
  author={Zhao, Tiancheng and Zhang, Tianqi and Zhu, Mingwei and Shen, Haozhan and Lee, Kyusong and Lu, Xiaopeng and Yin, Jianwei},
  journal={arXiv preprint arXiv:2207.00221},
  year={2022}
}

@article{hsieh2023sugarcrepe,
  title={Sugarcrepe: Fixing hackable benchmarks for vision-language compositionality},
  author={Hsieh, Cheng-Yu and Zhang, Jieyu and Ma, Zixian and Kembhavi, Aniruddha and Krishna, Ranjay},
  journal={Advances in neural information processing systems},
  volume={36},
  pages={31096--31116},
  year={2023}
}

@article{dumpala2024sugarcrepe++,
  title={Sugarcrepe++ dataset: Vision-language model sensitivity to semantic and lexical alterations},
  author={Dumpala, Sri Harsha and Jaiswal, Aman and Shama Sastry, Chandramouli and Milios, Evangelos and Oore, Sageev and Sajjad, Hassan},
  journal={Advances in Neural Information Processing Systems},
  volume={37},
  pages={17972--18018},
  year={2024}
}

@article{doveh2023dense,
  title={Dense and aligned captions (dac) promote compositional reasoning in vl models},
  author={Doveh, Sivan and Arbelle, Assaf and Harary, Sivan and Herzig, Roei and Kim, Donghyun and Cascante-Bonilla, Paola and Alfassy, Amit and Panda, Rameswar and Giryes, Raja and Feris, Rogerio and others},
  journal={Advances in Neural Information Processing Systems},
  volume={36},
  pages={76137--76150},
  year={2023}
}

@article{khan2024figclip,
  title={Figclip: Fine-grained clip adaptation via densely annotated videos},
  author={Khan, Zeeshan and Tapaswi, Makarand and others},
  journal={arXiv preprint arXiv:2401.07669},
  year={2024}
}

@article{basu2023augmenting,
  title={Augmenting clip with improved visio-linguistic reasoning},
  author={Basu, Samyadeep and Sanjabi, Maziar and Massiceti, Daniela and Hu, Shell Xu and Feizi, Soheil},
  journal={arXiv preprint arXiv:2307.09233},
  volume={2},
  number={6},
  pages={10},
  year={2023}
}

@inproceedings{sameni2024building,
  title={Building vision-language models on solid foundations with masked distillation},
  author={Sameni, Sepehr and Kafle, Kushal and Tan, Hao and Jenni, Simon},
  booktitle={Proceedings of the IEEE/CVF conference on computer vision and pattern recognition},
  pages={14216--14226},
  year={2024}
}

@inproceedings{zheng2024iterated,
  title={Iterated learning improves compositionality in large vision-language models},
  author={Zheng, Chenhao and Zhang, Jieyu and Kembhavi, Aniruddha and Krishna, Ranjay},
  booktitle={Proceedings of the IEEE/CVF Conference on Computer Vision and Pattern Recognition},
  pages={13785--13795},
  year={2024}
}

@article{yellinek20233vl,
  title={3vl: using trees to teach vision \& language models compositional concepts},
  author={Yellinek, Nir and Karlinsky, Leonid and Giryes, Raja},
  journal={arXiv preprint arXiv:2312.17345},
  volume={2},
  year={2023}
}

@inproceedings{herzig2023incorporating,
  title={Incorporating structured representations into pretrained vision \& language models using scene graphs},
  author={Herzig, Roei and Mendelson, Alon and Karlinsky, Leonid and Arbelle, Assaf and Feris, Rogerio and Darrell, Trevor and Globerson, Amir},
  booktitle={Proceedings of the 2023 Conference on Empirical Methods in Natural Language Processing},
  pages={14077--14098},
  year={2023}
}

@inproceedings{huang2024structure,
  title={Structure-clip: Towards scene graph knowledge to enhance multi-modal structured representations},
  author={Huang, Yufeng and Tang, Jiji and Chen, Zhuo and Zhang, Rongsheng and Zhang, Xinfeng and Chen, Weijie and Zhao, Zeng and Zhao, Zhou and Lv, Tangjie and Hu, Zhipeng and others},
  booktitle={Proceedings of the AAAI conference on artificial intelligence},
  volume={38},
  number={3},
  pages={2417--2425},
  year={2024}
}

@article{cascante2024natural,
  title={Natural language inference improves compositionality in vision-language models},
  author={Cascante-Bonilla, Paola and Hou, Yu and Cao, Yang Trista and Daum{\'e} III, Hal and Rudinger, Rachel},
  journal={arXiv preprint arXiv:2410.22315},
  year={2024}
}

@article{castro2024clove,
  title={Clove: Encoding compositional language in contrastive vision-language models},
  author={Castro, Santiago and Ziai, Amir and Saluja, Avneesh and Yuan, Zhuoning and Mihalcea, Rada},
  journal={arXiv preprint arXiv:2402.15021},
  year={2024}
}

@inproceedings{doveh2023teaching,
  title={Teaching structured vision \& language concepts to vision \& language models},
  author={Doveh, Sivan and Arbelle, Assaf and Harary, Sivan and Schwartz, Eli and Herzig, Roei and Giryes, Raja and Feris, Rogerio and Panda, Rameswar and Ullman, Shimon and Karlinsky, Leonid},
  booktitle={Proceedings of the IEEE/CVF Conference on Computer Vision and Pattern Recognition},
  pages={2657--2668},
  year={2023}
}

@inproceedings{zhang2024countercurate,
  title={Countercurate: Enhancing physical and semantic visio-linguistic compositional reasoning via counterfactual examples},
  author={Zhang, Jianrui and Cai, Mu and Xie, Tengyang and Lee, Yong Jae},
  booktitle={Findings of the Association for Computational Linguistics: ACL 2024},
  pages={15481--15495},
  year={2024}
}

@inproceedings{zhang2024contrasting,
  title={Contrasting intra-modal and ranking cross-modal hard negatives to enhance visio-linguistic compositional understanding},
  author={Zhang, Le and Awal, Rabiul and Agrawal, Aishwarya},
  booktitle={Proceedings of the IEEE/CVF Conference on Computer Vision and Pattern Recognition},
  pages={13774--13784},
  year={2024}
}

@article{bica2024improving,
  title={Improving fine-grained understanding in image-text pre-training},
  author={Bica, Ioana and Ili{\'c}, Anastasija and Bauer, Matthias and Erdogan, Goker and Bo{\v{s}}njak, Matko and Kaplanis, Christos and Gritsenko, Alexey A and Minderer, Matthias and Blundell, Charles and Pascanu, Razvan and others},
  journal={arXiv preprint arXiv:2401.09865},
  year={2024}
}

@inproceedings{kim2023misalign,
  title={Misalign, contrast then distill: Rethinking misalignments in language-image pre-training},
  author={Kim, Bumsoo and Jo, Yeonsik and Kim, Jinhyung and Kim, Seunghwan},
  booktitle={Proceedings of the IEEE/CVF international conference on computer vision},
  pages={2563--2572},
  year={2023}
}

@article{he2022generate,
  title={Generate, annotate, and learn: NLP with synthetic text},
  author={He, Xuanli and Nassar, Islam and Kiros, Jamie and Haffari, Gholamreza and Norouzi, Mohammad},
  journal={Transactions of the Association for Computational Linguistics},
  volume={10},
  pages={826--842},
  year={2022},
  publisher={MIT Press One Broadway, 12th Floor, Cambridge, Massachusetts 02142, USA~…}
}

@inproceedings{kumar2020data,
  title={Data augmentation using pre-trained transformer models},
  author={Kumar, Varun and Choudhary, Ashutosh and Cho, Eunah},
  booktitle={Proceedings of the 2nd workshop on life-long learning for spoken language systems},
  pages={18--26},
  year={2020}
}

@article{meng2022generating,
  title={Generating training data with language models: Towards zero-shot language understanding},
  author={Meng, Yu and Huang, Jiaxin and Zhang, Yu and Han, Jiawei},
  journal={Advances in Neural Information Processing Systems},
  volume={35},
  pages={462--477},
  year={2022}
}

@inproceedings{rosenberg2019speech,
  title={Speech recognition with augmented synthesized speech},
  author={Rosenberg, Andrew and Zhang, Yu and Ramabhadran, Bhuvana and Jia, Ye and Moreno, Pedro and Wu, Yonghui and Wu, Zelin},
  booktitle={2019 IEEE automatic speech recognition and understanding workshop (ASRU)},
  pages={996--1002},
  year={2019},
  organization={IEEE}
}

@article{varol2021synthetic,
  title={Synthetic humans for action recognition from unseen viewpoints},
  author={Varol, G{\"u}l and Laptev, Ivan and Schmid, Cordelia and Zisserman, Andrew},
  journal={International Journal of Computer Vision},
  volume={129},
  number={7},
  pages={2264--2287},
  year={2021},
  publisher={Springer}
}

@article{peng2017visda,
  title={Visda: The visual domain adaptation challenge},
  author={Peng, Xingchao and Usman, Ben and Kaushik, Neela and Hoffman, Judy and Wang, Dequan and Saenko, Kate},
  journal={arXiv preprint arXiv:1710.06924},
  year={2017}
}

@article{gao2022self,
  title={Self-guided noise-free data generation for efficient zero-shot learning},
  author={Gao, Jiahui and Pi, Renjie and Lin, Yong and Xu, Hang and Ye, Jiacheng and Wu, Zhiyong and Zhang, Weizhong and Liang, Xiaodan and Li, Zhenguo and Kong, Lingpeng},
  journal={arXiv preprint arXiv:2205.12679},
  year={2022}
}

@inproceedings{honovich2023unnatural,
  title={Unnatural instructions: Tuning language models with (almost) no human labor},
  author={Honovich, Or and Scialom, Thomas and Levy, Omer and Schick, Timo},
  booktitle={Proceedings of the 61st Annual Meeting of the Association for Computational Linguistics (Volume 1: Long Papers)},
  pages={14409--14428},
  year={2023}
}

@inproceedings{yang2022z,
  title={Z-lavi: Zero-shot language solver fueled by visual imagination},
  author={Yang, Yue and Yao, Wenlin and Zhang, Hongming and Wang, Xiaoyang and Yu, Dong and Chen, Jianshu},
  booktitle={Proceedings of the 2022 Conference on Empirical Methods in Natural Language Processing},
  pages={1186--1203},
  year={2022}
}

@inproceedings{zhu2023visualize,
  title={Visualize before you write: Imagination-guided open-ended text generation},
  author={Zhu, Wanrong and Yan, An and Lu, Yujie and Xu, Wenda and Wang, Xin and Eckstein, Miguel and Wang, William Yang},
  booktitle={Findings of the Association for Computational Linguistics: EACL 2023},
  pages={78--92},
  year={2023}
}

@article{azizi2023synthetic,
  title={Synthetic data from diffusion models improves imagenet classification},
  author={Azizi, Shekoofeh and Kornblith, Simon and Saharia, Chitwan and Norouzi, Mohammad and Fleet, David J},
  journal={arXiv preprint arXiv:2304.08466},
  year={2023}
}

@article{baradad2021learning,
  title={Learning to see by looking at noise},
  author={Baradad Jurjo, Manel and Wulff, Jonas and Wang, Tongzhou and Isola, Phillip and Torralba, Antonio},
  journal={Advances in Neural Information Processing Systems},
  volume={34},
  pages={2556--2569},
  year={2021}
}

@inproceedings{besnier2020dataset,
  title={This dataset does not exist: training models from generated images},
  author={Besnier, Victor and Jain, Himalaya and Bursuc, Andrei and Cord, Matthieu and P{\'e}rez, Patrick},
  booktitle={ICASSP 2020-2020 IEEE International Conference on Acoustics, Speech and Signal Processing (ICASSP)},
  pages={1--5},
  year={2020},
  organization={IEEE}
}

@article{he2022synthetic,
  title={Is synthetic data from generative models ready for image recognition?},
  author={He, Ruifei and Sun, Shuyang and Yu, Xin and Xue, Chuhui and Zhang, Wenqing and Torr, Philip and Bai, Song and Qi, Xiaojuan},
  journal={arXiv preprint arXiv:2210.07574},
  year={2022}
}

@inproceedings{hinterstoisser2018pre,
  title={On pre-trained image features and synthetic images for deep learning},
  author={Hinterstoisser, Stefan and Lepetit, Vincent and Wohlhart, Paul and Konolige, Kurt},
  booktitle={Proceedings of the European Conference on Computer Vision (ECCV) Workshops},
  pages={0--0},
  year={2018}
}

@inproceedings{li2021learning,
  title={Learning from noisy data with robust representation learning},
  author={Li, Junnan and Xiong, Caiming and Hoi, Steven CH},
  booktitle={Proceedings of the IEEE/CVF international conference on computer vision},
  pages={9485--9494},
  year={2021}
}

@inproceedings{li2019learning,
  title={Learning to learn from noisy labeled data},
  author={Li, Junnan and Wong, Yongkang and Zhao, Qi and Kankanhalli, Mohan S},
  booktitle={Proceedings of the IEEE/CVF conference on computer vision and pattern recognition},
  pages={5051--5059},
  year={2019}
}

@inproceedings{yao2020searching,
  title={Searching to exploit memorization effect in learning with noisy labels},
  author={Yao, Quanming and Yang, Hansi and Han, Bo and Niu, Gang and Kwok, James Tin-Yau},
  booktitle={International Conference on Machine Learning},
  pages={10789--10798},
  year={2020},
  organization={PMLR}
}

@inproceedings{liu2021noise,
  title={Noise-resistant deep metric learning with ranking-based instance selection},
  author={Liu, Chang and Yu, Han and Li, Boyang and Shen, Zhiqi and Gao, Zhanning and Ren, Peiran and Xie, Xuansong and Cui, Lizhen and Miao, Chunyan},
  booktitle={Proceedings of the IEEE/CVF Conference on Computer Vision and Pattern Recognition},
  pages={6811--6820},
  year={2021}
}

@inproceedings{ibrahimi2022learning,
  title={Learning with label noise for image retrieval by selecting interactions},
  author={Ibrahimi, Sarah and Sors, Arnaud and de Rezende, Rafael Sampaio and Clinchant, St{\'e}phane},
  booktitle={Proceedings of the IEEE/CVF Winter Conference on Applications of Computer Vision},
  pages={2181--2190},
  year={2022}
}

@article{wu2023brief,
  title={A brief overview of ChatGPT: The history, status quo and potential future development},
  author={Wu, Tianyu and He, Shizhu and Liu, Jingping and Sun, Siqi and Liu, Kang and Han, Qing-Long and Tang, Yang},
  journal={IEEE/CAA Journal of Automatica Sinica},
  volume={10},
  number={5},
  pages={1122--1136},
  year={2023},
  publisher={IEEE}
}

@article{wan2025wan,
  title={Wan: Open and advanced large-scale video generative models},
  author={Wan, Team and Wang, Ang and Ai, Baole and Wen, Bin and Mao, Chaojie and Xie, Chen-Wei and Chen, Di and Yu, Feiwu and Zhao, Haiming and Yang, Jianxiao and others},
  journal={arXiv preprint arXiv:2503.20314},
  year={2025}
}

@inproceedings{lim2025multi,
  title={Multi-granularity video object segmentation},
  author={Lim, Sangbeom and Kim, Seongchan and An, Seungjun and Cho, Seokju and Seo, Paul Hongsuck and Kim, Seungryong},
  booktitle={Proceedings of the AAAI Conference on Artificial Intelligence},
  volume={39},
  number={5},
  pages={5200--5208},
  year={2025}
}

@inproceedings{kirillov2023segment,
  title={Segment anything},
  author={Kirillov, Alexander and Mintun, Eric and Ravi, Nikhila and Mao, Hanzi and Rolland, Chloe and Gustafson, Laura and Xiao, Tete and Whitehead, Spencer and Berg, Alexander C and Lo, Wan-Yen and others},
  booktitle={Proceedings of the IEEE/CVF international conference on computer vision},
  pages={4015--4026},
  year={2023}
}

@article{wang2025internvideo2,
  title={Internvideo2. 5: Empowering video mllms with long and rich context modeling},
  author={Wang, Yi and Li, Xinhao and Yan, Ziang and He, Yinan and Yu, Jiashuo and Zeng, Xiangyu and Wang, Chenting and Ma, Changlian and Huang, Haian and Gao, Jianfei and others},
  journal={arXiv preprint arXiv:2501.12386},
  year={2025}
}

@inproceedings{wang2023towards,
  title={Towards Open-Vocabulary Video Instance Segmentation},
  author={Wang, Haochen and Yan, Cilin and Wang, Shuai and Jiang, Xiaolong and Tang, XU and Hu, Yao and Xie, Weidi and Gavves, Efstratios},
  booktitle={Proceedings of the IEEE/CVF International Conference on Computer Vision},
  year={2023}
}

@inproceedings{yang2019video,
  title={Video instance segmentation},
  author={Yang, Linjie and Fan, Yuchen and Xu, Ning},
  booktitle={Proceedings of the IEEE/CVF international conference on computer vision},
  pages={5188--5197},
  year={2019}
}

@article{hodson2022root,
  title={Root mean square error (RMSE) or mean absolute error (MAE): When to use them or not},
  author={Hodson, Timothy O},
  journal={Geoscientific Model Development Discussions},
  volume={2022},
  pages={1--10},
  year={2022},
  publisher={G{\"o}ttingen, Germany}
}

@inproceedings{kang2025clip,
  title={Is CLIP ideal? No. Can we fix it? Yes!},
  author={Kang, Raphi and Song, Yue and Gkioxari, Georgia and Perona, Pietro},
  booktitle={Proceedings of the IEEE/CVF International Conference on Computer Vision},
  pages={22436--22446},
  year={2025}
}

@article{choi2014text,
  title={Text analysis for detecting terrorism-related articles on the web},
  author={Choi, Dongjin and Ko, Byeongkyu and Kim, Heesun and Kim, Pankoo},
  journal={Journal of Network and Computer Applications},
  volume={38},
  pages={16--21},
  year={2014},
  publisher={Elsevier}
}

@misc{xu2024vta-ldm,  
      title={Video-to-Audio Generation with Hidden Alignment},   
      author={Manjie Xu and Chenxing Li and Yong Ren and Rilin Chen and Yu Gu and Wei Liang and Dong Yu},
      year={2024},
      eprint={2407.07464},
      archivePrefix={arXiv},
      url={https://arxiv.org/abs/2407.07464}, 
}
\end{document}